\documentclass{IEEEtran}
\usepackage{graphicx}
\usepackage{latexsym}
\usepackage{verbatim}
\usepackage{csquotes}
\usepackage{xcolor}
\usepackage{subcaption}
\usepackage{bm,amsfonts}
\usepackage{multirow}
\usepackage{xfrac}
\usepackage{comment,soul}

\begin{document}

\title{Towards General Game Representations:\\ Decomposing Games Pixels into Content and Style}

\author{\IEEEauthorblockN{Chintan Trivedi,
Konstantinos Makantasis, Antonios Liapis and
Georgios N. Yannakakis}
\IEEEauthorblockA{Institute of Digital Games,
University of Malta, 
Msida, Malta\\
Email: 
\{ctriv01, konstantinos.makantasis, antonios.liapis, georgios.yannakakis\}@um.edu.mt
}}
\maketitle

\begin{abstract}
On-screen game footage contains rich contextual information that players process when playing and experiencing a game. Learning pixel representations of games can benefit artificial intelligence across several downstream tasks including game-playing agents, procedural content generation, and player modelling. The generalizability of these methods, however, remains a challenge, as learned representations should ideally be shared across games with similar game mechanics. This could allow, for instance, game-playing agents trained on one game to perform well in similar games with no re-training. This paper explores how generalizable pre-trained computer vision encoders can be for such tasks, by decomposing the latent space into content embeddings and style embeddings. The goal is to minimize the domain gap between games of the same genre when it comes to game content critical for downstream tasks, and ignore differences in graphical style. We employ a pre-trained Vision Transformer encoder and a decomposition technique based on game genres to obtain separate content and style embeddings. Our findings show that the decomposed embeddings achieve style invariance across multiple games while still maintaining strong content extraction capabilities. We argue that the proposed decomposition of content and style offers better generalization capacities across game environments independently of the downstream task.
%
\end{abstract}

\section{Introduction}
\label{sec:Introduction}

\begin{figure}[!tb]
\centering
\includegraphics[width=\columnwidth]{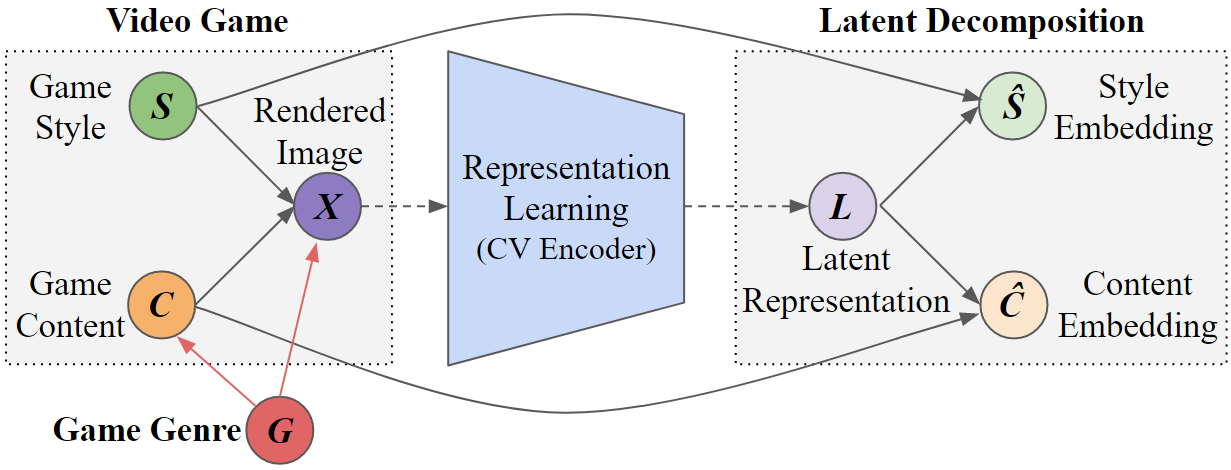}
\caption{We propose a causal relationship framework that views game pixels as causally dependent on style and content where the game genre only affects the content, but not the style. Solid lines represent a causal dependency and dashed lines represent flow of data. Our decomposition technique attempts to recover estimates of style and content from the latent representations of pre-trained computer vision encoders.}
\label{fig:causal_framework}
\end{figure}

Video game engines maintain an internal data representation of the game world capturing all necessary underlying variables describing the entities and objects existing within the virtual world \cite{rabin2005introduction,lewis2002game} such as the position of the player, the size of a room and the location of the opponent.
When this data representation is fed through the \emph{renderer} for generating the output image shown to the player it incorporates the visual styling of the game---defined in terms of e.g. textures and colors \cite{schodl2000video}. As a result, the visual \emph{style} and the game \emph{content}, represented as the internal game state, get entangled \cite{zhang2018separating} onto a high dimensional space (i.e. the RGB pixels of the game screen). This entanglement process is formally defined as \emph{causal dependency} and illustrated in Fig. \ref{fig:causal_framework}. 

\begin{figure*}[!tb]
    \centering
    \includegraphics[width=\linewidth]{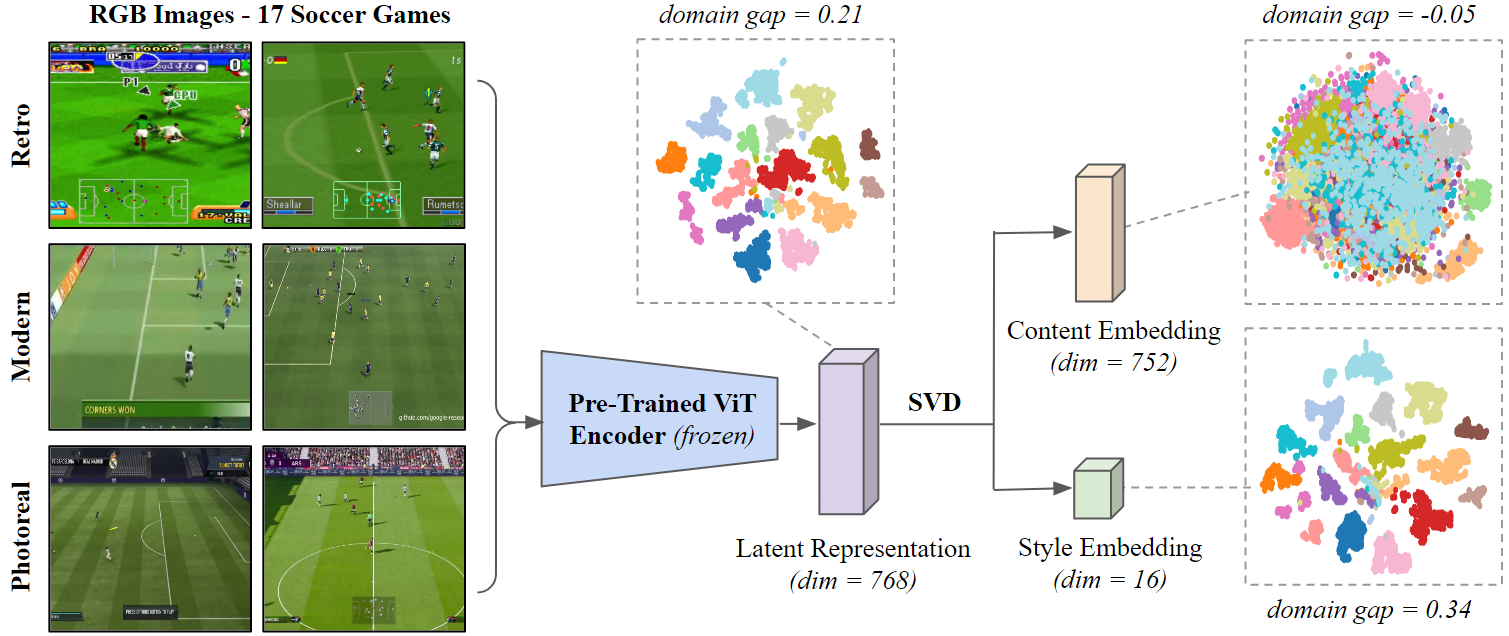}
    \caption{The proposed framework uses a dataset of multiple games from the same game genre (\emph{i.e.} 17 different soccer games in our example) to decompose the latent representation---based on singular value analysis with SVD---by learning transformations of the latent space to two lower-dimensional subspaces: one for \emph{content} and one for \emph{style}. In this example we use game footage frames from different soccer games (\emph{i.e.} same content) that vary in style (\emph{i.e.} retro, modern, photorealistic). The t-SNE plot visualizes the domain gap captured in the latent representations and shows how these differences are flattened into a separate style subspace which filters out the style gap. Content embeddings demonstrate virtually no domain gap among multiple soccer games. The dimensions of the derived embeddings for style and content in this example are, respectively, 16 and 752; 768 is the dimension of the ViT latent representation. Values next to each t-SNE plot represent the silhouette score of the examined embedding (which represents the domain gap); the lower the silhouette score, the smaller the gap.}
    \label{fig:domaingapexample}
\end{figure*}

When artificial intelligence (AI) methods are applied to games \cite{yannakakis2018artificial}, such internal data representations (e.g. the player's position) can be extracted directly from variables within the game engine \cite{barthet2021go, melhart2021affect,berner2019dota, silver2017mastering}. However, access to the game engine is a rarity and, hence, it severely limits the amount of games---especially commercial-standard games---available for game AI research. Even if one employs (or fine-tunes) pre-trained computer vision models for extracting latent representations from game pixels, the method will still not be generalizable and will suffer from what is known as the \emph{domain gap} problem \cite{wang2018deep} observed in latent representations of games (see Fig. \ref{fig:domaingapexample}). In particular, it has been shown \cite{trivedi2021contrastive} that graphical styling differences in games---arising from varying color palettes and abstract object designs---may cause a shift in the underlying distributions of latent representations as extracted from pre-trained models. 

Motivated by the lack of generic approaches that identify critical game state information (i.e. game content) solely through game frames, this paper introduces a method that processes the RGB pixels of the game footage and disentangles the game content from the game's style without relying on the game engine \cite{trivedi2022game}. Specifically, the introduced method learns to decompose content and style in a \emph{generalizable} zero-shot manner \cite{xian2018zero} for any game, thereby eliminating the technical challenges \cite{lecun2015deep} associated with fine-tuning neural network models. 

We propose a latent decomposition technique on the latent space of a pre-trained Vision Transformer \cite{dosovitskiy2020image} model, trained on the Imagenet dataset \cite{russakovsky2015imagenet} using the DINO self-supervised learning method \cite{caron2021emerging}. We show that it is possible to recover style and content embeddings from this latent space extracted from the game pixels without any further training required. Our primary contribution involves using a simple statistical method (i.e. singular value decomposition) to find latent directions \cite{harkonen2020ganspace,pham2022pca, tzelepis2021warpedganspace} that are unique across different games of the same genre, and define them as \emph{style}, while the remaining latent directions that are shared across these games are defined as \emph{content}.

To validate our hypotheses we test the method on two datasets. The Gen11 dataset introduced here contains 110k images from 11 game genres and 193 games that vary in style whereas the 3D-SSL dataset \cite{trivedi2022representations} contains game frames and corresponding internal game state information (e.g. player position) from 3 games of 3 difference genres. We conduct linear probing tasks \cite{anand2019unsupervised} on the derived content and style embeddings (using the 3D-SSL and Gen11 datasets, respectively) to recover relevant game state information (i.e. content) and graphical style of the game (e.g. retro, modern, photorealistic). Our findings suggest that by using the introduced method (see Fig. \ref{fig:domaingapexample}) it is possible to recover content information that exhibits style-invariance while maintaining efficient content extraction capacities in a generalizable fashion across game genres. 

The introduced method in this paper appears to be robust across all game genres tested and efficient at decomposing game footage aspects related to game content from those related to game style. The method is also generalizable to any game genre and directly applicable to any downstream task of AI in games.

\section{Background}
\label{sec:Background}


The number of studies with a focus on generalized representation learning within the domain of video games is limited to a few indicative papers. Trivedi \emph{et al.} \cite{trivedi2022representations} provide preliminary results with popular self-supervised learning (SSL) methods that are employed to learn accurate state representations of games. They showcase that SSL models are superior to pre-trained Imagenet models across three different 3D games: VizDoom \cite{kempka2016vizdoom}, the CARLA racing simulator \cite{dosovitskiy2017carla} and the Google Research Football Environment (GRFE) \cite{kurach2019google}. In \cite{trivedi2021contrastive} Trivedi \emph{et al.} train generalizable game models on 100k images across 175 games (and 10 game genres) and their findings suggest that contrastive learning is better suited for learning generalized game representations compared to conventional supervised learning. Lee \emph{et al.} \cite{lee2022multi} propose multi-game decision transformers which consider data from several Atari games as part of the training to make the encoder generalize across different Atari games. 

The advent of transferring advancements of self-attention \cite{vaswani2017attention} from the field of natural language processing to computer vision has enabled many advancements that derive from the inductive biases \cite{xu2021vitae} of the model architecture \cite{dosovitskiy2020image}. Caron \emph{et al.}  \cite{caron2021emerging} observe that Vision Transformers, when paired with self-supervised learning methods, are extremely well-suited for generalizing in comparison to convolutional networks. They note that the self-attention mechanism used in their ViT architecture can preserve spatial information in the images when compressing dimensionality for representation learning. Khan \emph{et al.} \cite{khan2021pretrained} build upon this finding and showcase that such pre-trained encoders are sufficient as backbone models to deploy in games. Their study, however, only supports that pre-trained models are sufficient for training agents on a particular game, and does not explore whether these trained agents can also be transferred to other games in a zero-shot manner.

\begin{figure*}[!tb]
    \centering
    \includegraphics[width=\textwidth]{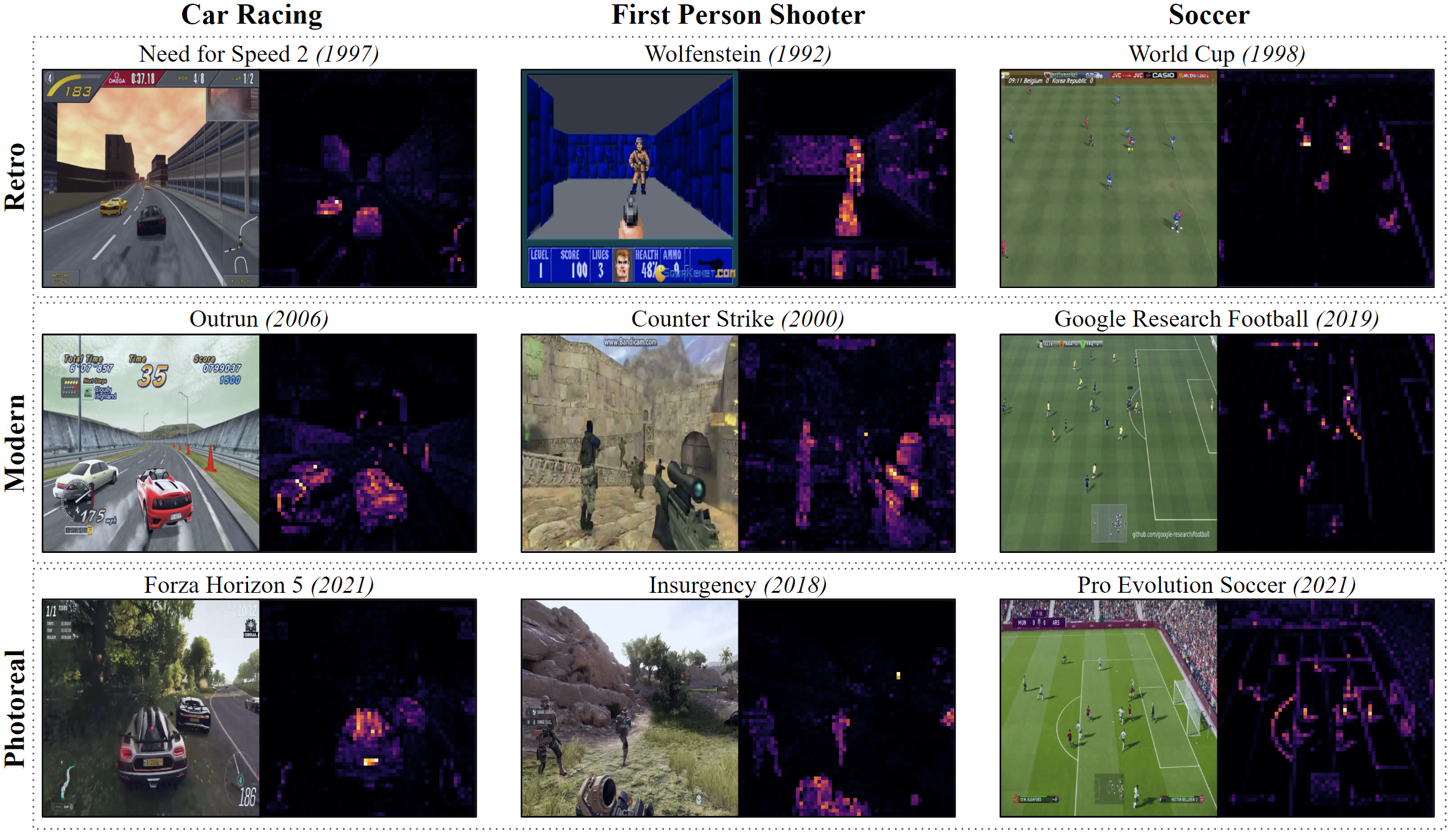}
    \caption{Visualizing the Spatial Attention Maps of DINO ViT-Base \cite{caron2021emerging} across three different game genres and graphical styles. The maps highlight the capabilities of self-attention layers \cite{dosovitskiy2020image} to extract content (e.g. car, enemy or player positions) while being invariant to the different game styles. The styles of retro, modern and photorealistic are shown in this example.}
    \label{fig:dino_attentions}
\end{figure*}

Computer vision studies that investigate the challenge of disentangling style from content of an image \cite{kotovenko2019content, kazemi2019style} are largely focused on the image style transfer paradigm \cite{jing2019neural}. When it comes to games, Trivedi \cite{trivedi2018style} showcased that it is possible to define style and content separation. 
Kim \emph{et al.} \cite{kim2021drivegan} proposed to learn style and content representations in a car driving simulator, defining content as spatial information and everything else as style. This style-content bifurcation \cite{liu2020measuring} cannot be applied to the definition of content across multiple games---even if they belong to the same genre---as we must cater for the game-specific differences in design choices and graphical fidelity of physical objects, especially those belonging to different game generations (e.g. low-bit retro games versus highly detailed modern games).

In contrast to all aforementioned studies, this paper aims to minimize the domain gap across games of the same genre. For that purpose we propose a novel decomposition technique on the latent representations of pre-trained encoder models by projecting them onto separate content and style subspaces. The method appears to be generic across different game genres, provided sufficient and representative data samples of game images.

\section{Style-Content Decomposition: Definitions}
\label{sec:Disentanglement}

In this section we discuss our proposed approach to style-content disentanglement from the perspective of video games. We first present our framework defining what we mean by \emph{style} and \emph{content} in video games, we then discuss our assumptions related to games and game genres, and finally we argue why existing pre-trained encoders are insufficient in terms of generalizability across games. This forms the basis of our proposed decomposition method discussed in Section \ref{sec:Methodology}.


Kim \emph{et al.} \cite{kim2021drivegan} define \emph{style} as information that does not depend on pixel location and \emph{content} as information that does. Within the car driving simulator used in their study, an example of style information would be weather condition while that of content would be placement of objects like buildings or cars in the rendered image of the scene. Inspired by these definitions, we follow similar terminology to refer to \emph{style} as the aesthetic design choice which influences visuals of the game, but has no bearing on the actual game world in its core elements related to gameplay dynamics. \emph{Content} instead can be viewed as any element of the game that influences the game-state existing within the predefined rules of the game world. These pre-defined rules are often shared across different games from the broader group of a game genre. 

Now, in contrast to \cite{kim2021drivegan}, since we propose working on multiple games from multiple game genres, we need to formalize what core elements of gameplay should be considered for all these games. For this, we propose a causal framework (see Fig. \ref{fig:causal_framework}) that utilizes the broader accepted notion of ``game genres'' to create a separation between what counts as content of a game and what gets interpreted as style. Figure \ref{fig:causal_framework} depicts a directed graph where a change in game genre only causes the content of the game to change, but not vice-versa. 
We hypothesize that a game may have an idiosyncratic visual style (e.g. skybox or wall textures). The content, however, is specified by conventions of the game genre: e.g. the types of game elements (ball, net), their visual representations (shape of a sniper rifle), their interactions (a soccer player may possess a ball by keeping it close), etc. In essence, our causal framework works under the assumption that different games from same genre share the same content space but not necessarily the same style.
An alternative way of looking at the differences between content and style within a game genre is their degree of subjectivity: the art style of a game can be viewed primarily as a subjective notion whereas content can be defined  objectively.


Based on our definition of style and content, we highlight the challenge of using pre-trained computer vision models for processing game pixels. We define \emph{domain gap} based on the framework defined earlier and quantify it as the influence of style in learning representations of different games belonging to the same genre. We therefore measure the domain gap in this paper based on the clustering quality metric \emph{silhouette score} \cite{rousseeuw1987silhouette, trivedi2021contrastive}, ranging between $-1$ and $+1$, which quantifies how clustered the representations of graphically different games of the same genre are in the embedding space. The higher the silhouette score, the larger the gap between clusters of representations of different games. When capturing content information that is assumed to share same information across graphically different games, we want this silhouette metric to be as low as possible with values close to 0 indicating no clear clusters. At the same time, the style information across different games should ideally demonstrate very high silhouette score values, with those being close to 1 indicating very well separated clusters. We utilize this definition of domain gap to claim whether a particular embedding space of a game genre demonstrates any domain gap or not.

Our hypothesis is that different games belonging to the same genre have the same content. This would mean that the latent representation of images of different games from the same genre would contain similar content. This would decrease the variance in the representations across games, and we can leverage this to disentangle style and content information from within the latent space of a pre-trained encoder. 

\section{Methodology}
\label{sec:Methodology}

The overall method we introduce in this paper is illustrated in Fig. \ref{fig:domaingapexample} through a soccer game genre example. We decompose the content of a game's image from its style, first, by employing a pre-trained encoder on the images that yields a latent representation for each frame. The details of the chosen Vision Transformer (ViT) \cite{dosovitskiy2020image} encoder are provided in Section \ref{sec:PixelEncoder}. Once the game footage frames from a number of different games (of the same genre) are encoded, they display a certain degree of domain gap expressed through the silhouette score of the t-SNE representations \cite{maaten2008visualizing} of the latent embedding (as in \cite{trivedi2022representations}). The latent representation is then decomposed into content and style via the use of the Singular Value Decomposition (SVD) method. The details of the SVD approach to content vs style decomposition are presented in Section \ref{sec:SVDonGames}. The section ends with a detailed description of the datasets used for our experiments (Section \ref{sec:Dataset}).


\subsection{Computer Vision Encoder}
\label{sec:PixelEncoder}

For all experiments reported in this paper, we use a pre-trained Vision Transformer (ViT) \cite{dosovitskiy2020image} backbone to extract latent representations of the game images. We use the ViT base encoder ($\sim85$ million parameters) with patch size 8 and pre-trained on Imagenet dataset \cite{russakovsky2015imagenet} employing the self-supervised DINO loss function \cite{caron2021emerging}. We also explored the more recent ViT Base distilled (patch size 14, knowledge distilled from larger $\sim1$ billion ViT model) from DINOv2 \cite{oquab2023dinov2}. Preliminary experiments, however, did not yield significant performance differences compared to DINO.

In Fig. \ref{fig:dino_attentions} we visualize the spatial self-attention maps used by the ViT architecture for 9 indicative games from our dataset. In particular, we obtain the attention activations from multi-head attention of the last transformer block, and average the activations across all 12 attention heads \cite{caron2021emerging}. This yields a 28x28 attention map, which we then flatten into an attention representation of dimension $784$. The attention maps depicted in Fig. \ref{fig:dino_attentions} showcase the capabilities of pre-trained self-attention layers to highlight a game's content while being invariant to the various graphical game styles.

\subsection{Decomposing Game Content and Style via SVD}
\label{sec:SVDonGames}

As mentioned earlier, the styles of different games belonging to the same genre can be substantially different. In contrast, their content---given that it is related to the game state---should not exhibit highly different patterns. We can verify this assumption by visualizing the attention heatmaps from within the encoder model, which make up a major part of a latent representation. These heatmaps attend to important spatial features in the image. Figure \ref{fig:dino_attentions} illustrated the effectiveness of the attention mechanism to highlight content-related visual features across different games while demonstrating robustness to the variant graphical style of the games. A preliminary analysis on the domain gap exhibited by these attention maps (see Table \ref{tab:domaingap}) further supports our assumption for all games examined. 

Based on the above assumption, we hypothesize that the latent image representation factors that exhibit high variation across data points are associated with the game style, while those that do not significantly vary are associated with game content. To validate this hypothesis, we first identify those {directions} in the latent representation space that capture the most variation in the data and use them as a feature representation of game style; we call these direction \emph{style components}. Then, we use the rest of the {directions} for representing content, which we call \emph{content components}.

For that purpose, we use the Singular Value Decomposition (SVD) \cite{wall2003singular} method to identify style components, which works as follows. Given a design matrix $\bm X\in \mathbb R^{N \times P}$ representing a dataset consisting of $N$ observation described by $P$ features, SVD will compute the decomposition $\bm X=\bm U \bm S \bm V^T$, where $\bm U$ is a unitary matrix, $\bm S$ is a diagonal matrix of singular values, and 
each of the columns of matrix $\bm V$ is associated with a \emph{singular value}. The SVD property we exploit in this study is that the variance in the data that a column of $\bm V$ can explain is proportional to the square of its associated singular value. Based on that property, we can use the $k$ columns of $\bm V$ associated with the $k$ largest singular values to encode style and the remaining columns to encode content. The parameter $k$ is selected empirically so that it maximizes the domain gap difference in games; \emph{i.e.} the difference in silhouette score values between the considered style representation and the derived content representation.

We should mention that popular dimensionality reduction methods such as the Principal Component Analysis (PCA) \cite{wall2003singular} can also be used as an alternative to SVD. Similarly to SVD, PCA finds those directions in the latent representation space that capture most of the variance in the data. PCA, however, operates on the covariance matrix of the centered design matrix (\emph{i.e.} the mean of every column is equal to zero). To avoid such data transformations, we use SVD in this study. Our experimental results did not indicate any significant difference between the two approaches and we, thus, employ SVD for all experiments reported here.

\begin{figure*}[!tb]
    \centering
     \includegraphics[width=\linewidth]{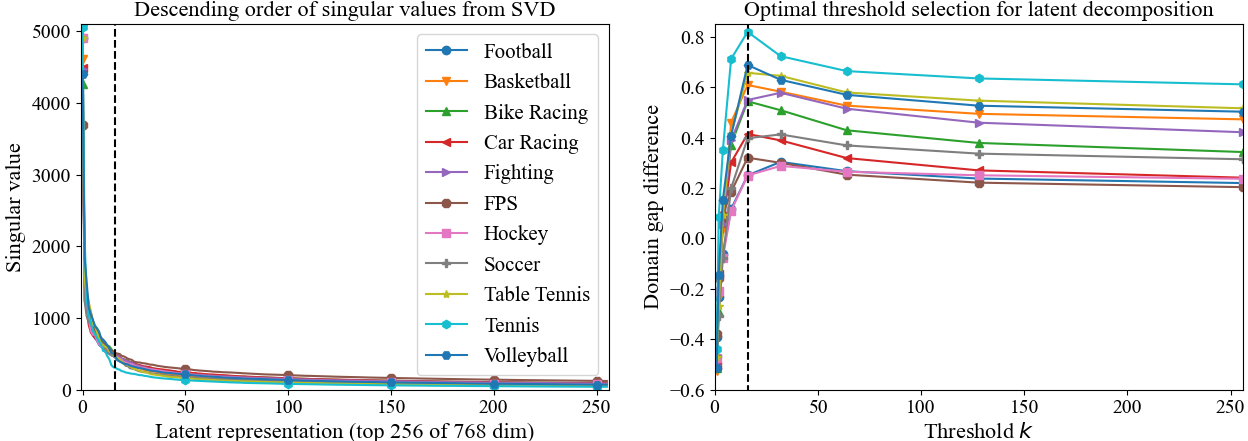}
     \caption{Decomposing to content and style with SVD. Left: top 256 singular values of SVD across all 768 dimensions of latent representation for all 11 game genres examined. Right: determining the optimal $k$ per game genre that maximizes the domain-gap difference (\emph{i.e.} difference in silhouette score values between content and style representations). The domain-gap difference is visualized for the following values of $k$: 1, 2, 4, 8, 16, 32, 64, 128 and 256. The dotted line in both graphs depicts the determined decomposition point between style (first 16 dimensions) and content (remaining 752 dimensions).}
     \label{fig:svd}
\end{figure*}


\subsection{Datasets}
\label{sec:Dataset}

Our proposed framework requires a specific dataset structure, where it is possible to categorize a game under a single game genre. For this purpose, we select the \emph{Sports10} dataset introduced by Trivedi \emph{et al.} \cite{trivedi2021contrastive}, which contains gameplay images from 175 games across 10 sports game genres with varying graphical styling. In addition to Sports10 games, we consider data from the first-person shooter (FPS) genre. In particular, we consider $10,000$ images of 18 different FPS games, 6 each across three graphical stylings, similar to the format of other game genres in the {Sports10} dataset.

\begin{table}
\begin{center}
\caption{Summary of the video game datasets used in this study.}\label{tab:datasets}
\begin{tabular}{|l||l|l|}
\hline
\textbf{Dataset Name} & Gen11 & 3D-SSL \\ \hline \hline
\textbf{Game Genres}  & \begin{tabular}[c]{@{}l@{}}Football, Basketball,\\  Bike Racing, Car Racing,\\  Fighting, FPS, Hockey, \\ Soccer, Table Tennis,\\ Tennis and Volleyball\end{tabular} & \begin{tabular}[c]{@{}l@{}}Car Racing, FPS, \\Soccer\end{tabular} \\ \hline
\textbf{Total Games}  & 193 & 3 \\ \hline
\textbf{Total Images} & 110,000 & 40,608 \\ \hline
\textbf{Labels} & \begin{tabular}[c]{@{}l@{}}Graphics Styling (Retro, \\ Modern, Photoreal)\end{tabular} & \begin{tabular}[c]{@{}l@{}}Internal Game State \\ (Positions of Player, \\ Opponents and other\\ Game Objects)
\end{tabular} \\ \hline
\textbf{Usage} & \begin{tabular}[c]{@{}l@{}}Latent Decomposition,  \\ Style Prediction 
\end{tabular} & Content Extraction \\ \hline
\end{tabular}
\end{center}
\end{table}

The combined Gen11 dataset (as we name it here) contains 193 games across 11 game genres in total. In terms of graphical styling, it contains 58 games labeled as \enquote{retro}, 90 as \enquote{modern} and 45 as \enquote{photorealistic}. The detailed list of games used and their properties are provided in Table \ref{tab:datasets}. We use this dataset in our study for two experiments. First, we attempt to quantify the domain gap across 11 different game genres and to learn a projection matrix for latent decomposition, as described in Section \ref{sec:Disentanglement}. Second, we use the three style labels of all games to evaluate our style embeddings as detailed in Section \ref{subsec:StyleEmbeddings}.

In addition to Gen11, we also use the 3D-SSL dataset introduced in  \cite{trivedi2022representations} for evaluating content embeddings (see Section \ref{subsec:ContentEmbeddings}). This dataset contains paired images and game states obtained for three games by interfacing with the game engine to extract exact variables that describe the state of a game at any given instance. The properties of the 3D-SSL dataset are outlined in Table \ref{tab:datasets}.

\section{Results}
\label{sec:Results}

This Section presents the decomposition analysis of the game genres considered (Section \ref{subsec:DecomposingLatent}), and assesses the embeddings of content (Section \ref{subsec:ContentEmbeddings}) and style (Section \ref{subsec:StyleEmbeddings}) derived through our method.

\subsection{Decomposing Latent into Content and Style}
\label{subsec:DecomposingLatent}

\begin{table}
\centering
\caption{Domain gap for \textbf{style} embeddings (higher is better) on Gen11 dataset for different selection strategies with $k=16$. We quantify domain gap as the silhouette score of the examined embedding.}
\begin{tabular}{|l|c|c|c|c|}
\hline
\textbf{Genre} & \textbf{Top $k$} & \textbf{Random $k$} & \textbf{Last $k$} & \begin{tabular}[c]{@{}c@{}}\textbf{Top \sfrac{k}{2} +}\\ \textbf{Random \sfrac{k}{2}} 
\end{tabular} \\ \hline  \hline
Football & 0.240&-0.179&-0.087&0.171 \\ 
Basketball & 0.517	&	-0.150	&	-0.211	&	0.468 \\ 
Bike Racing & 0.495	&	-0.116	&	-0.195	&	0.413 \\ 
Car Racing & 0.382	&	-0.075	&	-0.195	&	0.343 \\ 
Fighting & 0.396	&	-0.241	&	-0.259	&	0.321 \\ 
FPS & 0.261	&	-0.138	&	-0.183	&	0.184 \\ 
Hockey & 0.219	&	-0.167	&	-0.109	&	0.148 \\ 
Soccer & 0.348	&	-0.111	&	-0.195	&	0.269 \\ 
Table Tennis & 0.523	&	-0.172	&	-0.191	&	0.475 \\ 
Tennis & 0.677	&	-0.162	&	-0.226	&	0.698 \\ 
Volleyball & 0.488	&	-0.262	&	-0.179	&	0.426 \\ \hline
\end{tabular}
\label{tab:topk}
\end{table}

To assess the decomposition method we introduce in this paper, we explore the results of SVD and choose appropriate directions from its outputs. We apply SVD on the attention representation ($D=768$) of each game genre (10k images) and visualize the highest 256 singular values (out of 768) in the left graph of Fig. \ref{fig:svd}. We observe that most of the variance in the representations of different games within a genre comes from very few dimensions of the latent; this cut-off point (threshold $k$) is highlighted using the vertical dashed line after first few dimensions. 
The \enquote{flat} nature of this graph to the right of the dashed line indicates there is plenty of information in the latent representations that remains consistent across games despite having different graphical styling in its input. An encoder that is not robust to such variations would have a more gradual \enquote{decline} for the singular values. To verify our assumption that the selection of \enquote{top} $k$ singular values projects the data into a subspace that captures most variance, we perform a sanity check by comparing it against \enquote{random $k$}, \enquote{last $k$} and \enquote{top \sfrac{k}{2} + random \sfrac{k}{2}} strategies presented in Table \ref{tab:topk}. As expected, selecting singular values from the top of the descending order (refer Fig. \ref{fig:svd}) provides highest domain gap on average, ensuring that SVD is an effective method for selecting high variance dimensions of the latent.

\begin{table}
\centering
\caption{Domain gap (as silhouette score) for different embeddings and game genres. Lower values indicate smaller domain gap. \label{tab:domaingap}}
\begin{tabular}{|l|c|c|c|c|c|}
\hline 
\textbf{Genre} & \textbf{RandViT} & \textbf{Latent} & \textbf{Attention} & \textbf{Content} & \textbf{Style} \\
\hline \hline

Football & 0.147 & $0.138$ & $-0.015$ & $-0.010$ & $0.240$ \\
Basketball & 0.201 & $0.290$ & $-0.012$ & $-0.091$ & $0.517$ \\
Bike Racing & 0.093 & $0.236$ & $0.021$ & $-0.048$ & $0.495$ \\
Car Racing & -0.137 & $0.161$ & $0.056$ & $-0.032$ & $0.382$ \\
Fighting & 0.047 & $0.218$ & $-0.010$ & $-0.153$ & $0.396$ \\
FPS & -0.144 & $0.104$ & $0.026$ & $-0.059$ & $0.261$ \\
Hockey & 0.030 & $0.140$ & $-0.096$ & $-0.030$ & $0.219$ \\
Soccer & 0.353 & $0.213$ & $-0.032$ & $-0.050$ & $0.348$ \\
Table Tennis & 0.317 & $0.340$ & $0.073$ & $-0.133$ & $0.523$ \\
Tennis & 0.626 & $0.441$ & $0.022$ & $-0.142$ & $0.677$ \\
Volleyball & 0.167 & $0.276$ & $-0.007$ & $-0.199$ & $0.488$ \\
\hline 
\end{tabular}
\end{table}

Next, we need to assess the optimal value for the cut-off $k$ based on a fixed criterion. The graph on the right of Figure \ref{fig:svd} depicts the method we propose for selecting the optimal value of threshold $k$. Here, for a given cutoff value $k$, we measure the domain gap (lower is better for content) observed in the content embedding (dimension is $D-k$) and subtract it from the domain gap (higher is better for style) observed in style embedding (dimension is $k$). The optimal threshold for 9 out of 11 game genres is $k=16$. For the game genres of Football and Hockey, the optimal $k$ value turns out to be 32. However, the difference of silhouette scores between style and content representations at $k=16$ for these two game genres is not noticeably lower (between 8\% and 10\%). Hence, for the sake of consistency, we select $k=16$ as a general cutoff point for separating the style and content embeddings across all game genres. This analysis yields style embeddings of dimensionality $k=16$ and content embeddings of dimensionality $D-k=752$.

Table \ref{tab:domaingap} shows an analysis in terms of the domain gap for our style and content embeddings for every game genre. Table \ref{tab:domaingap} validates our assumption that the top $k$ singular values of SVD are representative of style whereas the bottom $D-k$ are representative of content. We still need to verify whether these embeddings capture the notions of game \emph{style} and game \emph{content} as defined in Section \ref{sec:Disentanglement}. For this purpose we run a linear probing analysis on both derived embeddings to predict elements of style versus features of content as detailed below.

\subsection{Evaluating Content Embeddings} \label{subsec:ContentEmbeddings}

We test whether the content embeddings we derived earlier are (a) invariant to the style of different games of the same genre and (b) representative of accurate game state information. We use the 3D-SSL dataset \cite{trivedi2022representations} to evaluate to what extent our content embeddings accurately capture game state information. 

In particular, we follow the linear probing protocol introduced in \cite{anand2019unsupervised} and run simple linear regression on game-state variables following the protocol of \cite{trivedi2022representations}. Indicative game-state variables for the three games of the 3D-SSL dataset \cite{trivedi2022representations} include the position of the players and the ball (94 variables for GRFE), the position of enemies (12 variables for VizDoom) and the position of the car and nearby traffic (7 variables for CARLA). We test four embeddings: (a) the latent representation of the pre-trained ViT (dim=768), which represents the \emph{upper bound} in terms of game state prediction potential for this experiment; (b) the latent representation (dim=768) of a randomly initialized ViT (RandViT) as a baseline to assess the impact of the training process; (c) the attention map (dim=784) of the pre-trained ViT, to assess whether features that only contain spatial information suffice for this task; (d) the content embeddings (dim=752) derived from our method. The performance of the linear probe is based on the coefficient of determination (commonly known as $R^2$ score) of this model, averaged across all game state variables.

\begin{figure}[!tb]
    \centering
    \includegraphics[width=0.48\textwidth]{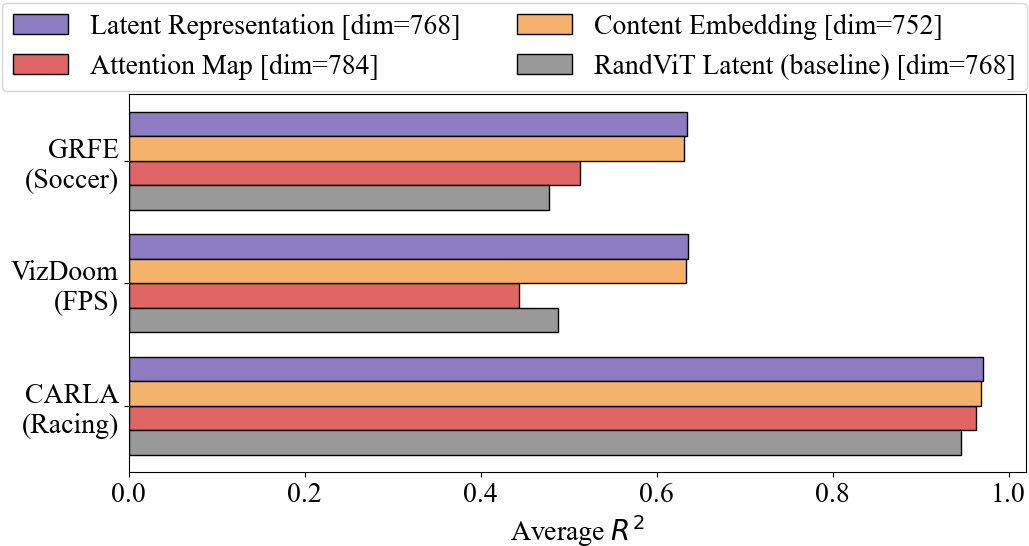}
    \caption{Linear probing results, measured as $R^2$ score averaged across all game state variables of the 3D-SSL dataset \cite{trivedi2022representations}.
    \label{fig:linear_regression}}
\end{figure}

Figure \ref{fig:linear_regression} presents the key results of this experiment. As expected, the full latent representation of the pre-trained ViT reaches the highest $R^2$ scores, although it is more capable of aligning with values for CARLA than in the other two games. Moreover, the randomly initialized ViT performs poorly, with a relative decrease of 24\%, 23\% and 2.5\% for GRFE, VizDoom, CARLA respectively. Evidently the CARLA dataset is fairly easy to predict, as even untrained models can perform relatively well. The attention layer underperforms as well, with a relative decrease of 19\%, 30\%, 0.8\% for GRFE, VizDoom, CARLA respectively. It seems while attention is good at detecting objects on the screen, it is not sufficient at distinguishing between different object types which would require additional visual information. This is not surprising, considering that games also require identifying the detected objects, for example differentiating health packs from enemies in shooters or players of different teams and the ball in soccer. Finally, it is evident that content embeddings contain most of the information needed for these tasks, reaching $R^2$ scores very close to the upper bound (relative decrease from the full latent representation by 0.46\%, 0.43\% and 0.23\% for GRFE, VizDoom, CARLA respectively). This showcases the robustness of the derived content embedding to capture key game state information at par with the latent representations. 
%
%
%

As shown in Table \ref{tab:domaingap}, content embeddings can bridge the domain gap, especially compared to the latent representations across all 11 game genres (Gen11 dataset). We observe that the derived content embeddings are at par with the latent representations at predicting critical game state information. In comparison, the attention maps only capture spatial information from the pixels of the game footage and are less accurate at predicting the state of the game. It is also important to note that the dimensionality of the content embeddings is smaller (\emph{i.e.} 752) than that of the attention embeddings (\emph{i.e.} 784) and the latent representation (\emph{i.e.} 768). 



\subsection{Evaluating Style Embeddings}
\label{subsec:StyleEmbeddings}

\begin{figure}[!tb]
    \centering
    \includegraphics[width=0.48\textwidth]{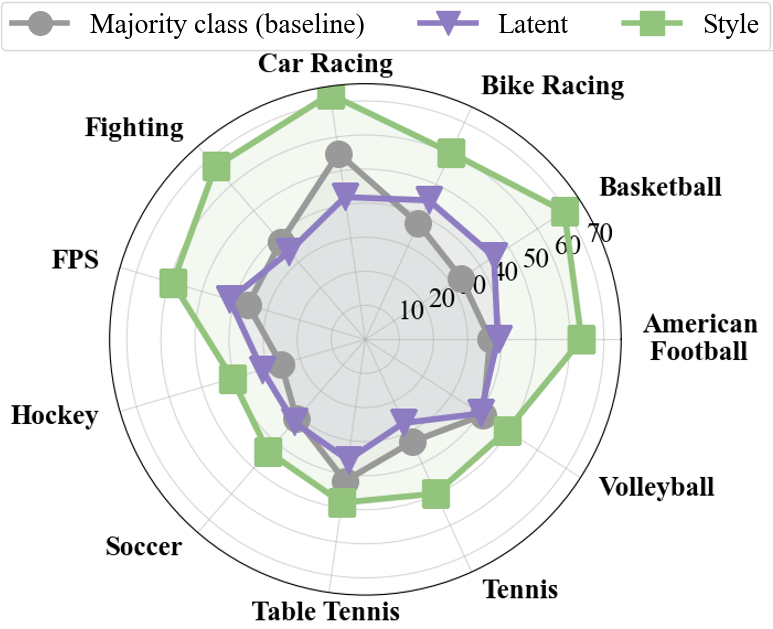}
    \caption{Classification accuracy of the game's style (retro, modern, photorealistic) across 11 genres. We compare the latent representation (Latent) against the style embedding (Style) as input representations for the classifier. The majority class is also depicted (Baseline). Results are averaged across 10 folds, where 3 games per genre are selected at random to form the test set.} 
    \label{fig:linear_classification}
\end{figure}

In this section we test the capacity of the derived style embeddings to recover descriptive information about a game's graphic style in terms of retro, modern and photorealistic graphics. We use a similar linear probing technique as in Section \ref{subsec:ContentEmbeddings}, but this time we learn to predict the graphics style class (\emph{i.e.} retro, modern and photorealistic) as a single-output classification task. Figure \ref{fig:linear_classification} shows classification accuracies for test sets obtained using the leave-three-games-out cross-validation scheme (\emph{i.e.} one game from each style label). This bifurcation of the dataset into training and evaluation sets is performed randomly over 10 folds, averaged and depicted in Fig. \ref{fig:linear_classification}. A majority vote classifier is used as our baseline for this dataset, calculated on the training set in every fold.

For all 11 game genres, we notice that style embeddings are able to recover graphical styling information with higher accuracy than the baseline, with test accuracies as high as 72\%. For comparison, we also test the capacity of the latent representations as input to this linear classifier. Despite the latent representations integrating style information, its classification accuracy is not able to reach that of the style embeddings; with fewer features (16 versus 768) classification is easier for style embeddings than the full latent representation. The style prediction results highlighted in Fig. \ref{fig:linear_classification}, combined with domain gap results summarized in Table \ref{tab:domaingap}, collectively provide strong evidence that the derived style embeddings are meaningful representations and efficient predictors of style across all game genres examined. 

\section{Discussion}
\label{sec:Discussion}

Our findings provide evidence that only a few features of the latent representation of a pre-trained Visual Transformer account for the style of a game, with remaining ones accounting for content information that can be re-used across other games in the same genre. By testing both style embeddings and content embeddings in fairly simple tasks (predicting graphical style and game variables respectively), we offer evidence that the split is a viable way forward for more complex downstream tasks. Indicatively, a more complex downstream task that consider game footage pixels is affect modelling \cite{makantasis2019pixels}, where self-reported arousal or valence can be predicted from the gameplay context alone. In such a case, training models on top of only the content features in human data reported in one game would allow us to reuse this predictive model on other games of the same genre. However, if our affect prediction task has a dataset with multiple games available in the training set---as in \cite{melhart2021affect}---then using a disentangled style embedding in addition to the generalized content embedding can help us build a global model of style for the entire game genre. Such a model can enhance its predictive capacity of player affect that may relate to the game style as well on top of the content information. Ultimately, how to best use the disentangled content and style embeddings for improving generalizability will vary according to the downstream task. Beyond the simple classifiers tested in this paper, in future studies we plan to test the transferability of the derived embeddings across games for a number of downstream tasks, including game-playing agents, player modeling and content generation.

In this study, we leverage SVD for splitting ViT latent representations into style and content embeddings. The core advantages of the SVD method proposed is its simplicity and effectiveness. To obtain the content and style embeddings successfully we are neither required to retrain or finetune the ViT encoder, nor do we need to train any additional linear heads on top of the latent representation. In addition to computational resources, we also minimize the complexity of using stochastic learning techniques through the use of simpler statistical methods like SVD or PCA. However, future work could explore more nuanced ways of deriving style versus content embeddings, possibly using the same methodology of Sections \ref{subsec:ContentEmbeddings} and \ref{subsec:StyleEmbeddings} to validate their efficiency.

It is worth noting that our methodology pre-supposes that any game fits neatly into a specific game genre. In our experiments, we used a dataset of 11 game genres with clear-cut boundaries (a soccer game can not be a volleyball game) imposed by real-life sports. First-person shooters are not tied to a specific real-life sport but tend to be rather clearly defined as well. However, other game genres have less clear boundaries, such as action role-playing games, and may include very different games (which may also overlap with other genres). We hypothesize that visual style in such games will be more difficult to disentangle from content, although our current method does not juxtapose games from different game genres. More importantly, the visual depiction of game genres in our study (e.g. soccer) is visually close to real-life sports visualizations or camera-perspectives (for FPS); therefore, the ViT pre-trained on ImageNet data can capture those easily. For games where the representation is not as close to the real-world (or with little ambient light, as in horror games), the pre-trained ViT model might not suffice for extracting content through self-attention as shown in Fig.~\ref{fig:dino_attentions}. Future work should explore which granularity in defining ``genres'' is necessary, and how far pre-trained models can be of use in a broader set of games.

\section{Conclusion}
\label{sec:Conclusion}

This paper focuses on generalizable representation learning within the domain of games. We present a novel framework that is able to model game pixels as separate style and content embeddings for different game genres. This disentanglement allows for deep learning models based on shared content representations across multiple games of same genre, treating style as a game-specific property. Our findings suggest that the derived content and style embeddings capture, respectively, critical elements of the game state (\emph{e.g.} position of a player) and the game's graphical styling (\emph{e.g.} retro vs photorealistic). Additionally, the clustering quality analysis also emphasize the effectiveness of these embeddings in reducing the domain gap across different games of same genre. Based on the experiments performed in this paper we can argue that the introduced method yields more generalizable and reusable game models, both for pixels to latent embeddings (representation learning) and for latent embeddings to pixels (generative modeling). We hope that the paper will spur further research in downstream applications with zero-shot transferability across games such as game-playing agents, player modeling and procedural content generation.


\bibliographystyle{IEEEtran}
\bibliography{ecai}

\end{document}